\documentclass[11pt]{article}
\usepackage{INTERSPEECH2019}
\usepackage{multirow}
\usepackage{rotating}
\usepackage{hyperref}

\usepackage{graphicx}
\usepackage{amssymb,amsmath,bm}
\usepackage{textcomp}
\usepackage{booktabs}
\usepackage[textfont=it,tableposition=top]{caption}
\usepackage{siunitx}
\usepackage{xspace}
\usepackage{url}
\usepackage{lipsum}  
\usepackage{tipa}
\usepackage{enumitem}

\title{Robust Open-Set Spoken Language Identification and the CU MultiLang Dataset\ \\ \small{\href{https://www.recotechnologies.com}{Recognition Technologies, Inc.}}\\
  \small{Technical Report: \href{https://www.recognitiontechnologies.com/~beigi/ps/RTI-20230828-01.pdf}{RTI-20230828-01}}\\
  \small{\href{http://dx.doi.org/10.13140/RG.2.2.22716.21122}{DOI: 10.13140/RG.2.2.22716.21122}\vspace{-0.3cm}}
}
\name{Mustafa Eyceoz$^1$, Justin Lee$^1$, Siddharth Pittie$^1$, Homayoon Beigi$^{1,2}$}
\address{
  $^1$Columbia University, New York, USA\\
  $^2$Recognition Technologies, Inc., New York, USA}
\email{me2680@columbia.edu, jjl2245@columbia.edu, sp4013@columbia.edu, beigi@recotechnologies.com}

\begin{document}

\maketitle

\begin{abstract}

Most state-of-the-art spoken language identification models are closed-set; in other words, they can only output a language label from the set of classes they were trained on. Open-set spoken language identification systems, however, gain the ability to detect when an input exhibits none of the original languages. In this paper, we implement a novel approach to open-set spoken language identification that uses MFCC and pitch features, a TDNN model to extract meaningful feature embeddings, confidence thresholding on softmax outputs, and LDA and pLDA for learning to classify new unknown languages. We present a spoken language identification system that achieves 91.76\% accuracy on trained languages and has the capability to adapt to unknown languages on the fly. To that end, we also built the CU MultiLang Dataset, a large and diverse multilingual speech corpus which was used to train and evaluate our system. 

\end{abstract}
\noindent\textbf{Index Terms}: Spoken language identification, open-set, closed-set, MFCC, TDNN, softmax, threshold, LDA, pLDA, dataset

\section{Introduction}

Spoken Language Identification is the process of determining the language being spoken from an input audio. There are two subdivisions to the language identification (LID) problem: open-set and closed-set. 
In closed-set LID, the set of languages to identify is predefined, and for
every audio input, the ``most probable'' language within the set is
outputted. However in open-set LID, there is the option to ``reject'' that prediction and detect when the audio input matches none of the known languages well. 

In this paper, we improve upon a modern approach to the open-set spoken LID problem that utilizes Mel-frequency cepstral coefficients (MFCC) and pitch
features~\cite{r:beigi-sr-book-2011} with a Time-Delay Neural Network (TDNN)~\cite{r-m:waibel-1989,r-m:peddinti-2015} and softmax output to reject
an input audio and label it as an ``unkown'' language. Our TDNN described in this paper is architected and trained from scratch without any pre-trained models or transfer learning. 

To improve on that system,
our first step was the construction of the CU MultiLang dataset: an open-source speech dataset containing 51 languages with at most 10 hours of speech data for each language with corresponding transcriptions. 
The languages were chosen to cover a diverse array of language families, and the samples themselves were randomly chosen to maximize speaker diversity. 
We then selected 32 languages as the in-set base, with the rest being out-of-set.


The next improvement was the addition of Linear Discriminant Analysis (LDA) and Probabilistic LDA (pLDA)~\cite{r:beigi-sr-book-2011, Izenman2008, 10.1007/11744085_41} to the system architecture, which are used to perform efficient classification of out-of-set languages. 
By taking the output of the TDNN when
peeling back its last two layers, the system obtains a meaningful feature embedding we will call the ``language representation vector'' after which it performs LDA and pLDA 
to match the input with one of the previously learned out-of-set languages. 
The LDA and pLDA allow the system to learn new languages on the fly: when encountering an input that the system determines to be from an out-of-set language, instead of retraining the TDNN to learn it, the language representation vector from our diversely
trained and generalized TDNN is used it to refit our LDA
and pLDA, easily adding a new language class.

\section{Related work}

In ``Modernizing Open-Set Speech Language
Identification''~\cite{oldpaper}, Eyceoz et al. built a system that extracted MFCC and pitch
features, passed them through a TDNN with softmax
output, and achieved a 95\% in-set language classification accuracy,
trained and tested on only seven languages. A number of static confidence thresholds were tested for the detection of
out-of-set languages. Two additional languages were designated as out-of-set languages and the system achieved an out-of-set detection accuracy of 80.4\%.
Our open-set spoken LID system of this paper improves upon the aforementioned approach.

\section{The CU MultiLang Dataset}

In order to build a robust and generalizable system, we created the the CU MultiLang Dataset by pulling from open-source datasets on OpenSLR~\cite{openslr}, VoxForge~\cite{Voxforge}, VoxLingua107~\cite{VoxLingua}, and a number of other sources to compose a 51-language speech data
corpus with over 400 hours worth of samples. Each language has a
selection of utterances that sum up to at most 10 hours of speech data, with each utterance having an accompanying text
transcription. The individual utterances were specifically selected in
an automated fashion to maximize speaker and dialect diversity within
each language. Similarly, the languages were also carefully chosen to cover a large number of diverse language families.
The dataset is publicly available at \path{https://www.speechdata.com/datasets/cu_multilang}

For Romance languages, the CU MultiLang Dataset contains French of various
regions from MediaSpeech~\cite{mediaspeech2021} and
African Accented French Corpus~\cite{AAFrench}, Spanish of various
regions from MediaSpeech and Crowdsourcing Latin American
Spanish Dataset~\cite{guevara-rukoz-etal-2020-crowdsourcing}, and
Romanian and Italian from VoxLingua107. For Germanic languages, it contains English from Free ST American English
Corpus~\cite{freest_english}, German from Swiss Parliaments
Corpus~\cite{swissparliament}, Icelandic from
Samromur~\cite{mollberg-etal-2020-samromur}, Norwegian and Swedish
from VoxLingua107, and Dutch from VoxForge. For Semitic languages,
it contains Arabic of various regions from
MediaSpeech and Hebrew from VoxForge. For Slavic languages, it contains Russian from Russian LibriSpeech
Dataset~\cite{russian_librispeech}, and Ukrainian, Croatian, and
Bulgarian from VoxForge. For Indo-Iranian languages, it
contains Persian from VoxForge, Kashmiri from Kashmiri Data
Corpus~\cite{kashmiri}, Pashto from VoxLingua107, and Bengali from Large Bengali ASR Training Dataset~\cite{kjartansson-etal-sltu2018}.
For other Indo-European languages, it contains Greek and
Albanian from VoxForge, Armenian from VoxLingua107, and Catalan from Crowdsourced high-quality Catalan speech Dataset~\cite{kjartansson-etal-2020-open}. For Dravidian languages, it contains Tamil, Malayalam, and Telegu all from Crowdsourced High-quality Multi-speaker Speech
Dataset~\cite{he-etal-2020-open}. For Sino-Tibetan languages, it contains Mandarin and Tibetan both from VoxLingua107, and
Burmese from Crowdsourced High-quality Burmese Speech
Dataset~\cite{oo-etal-2020-burmese}. For Austro-Tai languages, it contains Javanese from Large Javanese ASR Training
Dataset~\cite{kjartansson-etal-sltu2018} and Iban from Iban
Dataset~\cite{Juan14}. For Altaic languages, it contains
Turkish from MediaSpeech, Japanese from VoxLingua107, Korean from
Zeroth-Korean~\cite{Zeroth-Korean}, and Uyghur from
THUYG-20~\cite{thuyg}. For Niger-Congo languages, it
contains Ewe, Hausa, Lingala, Yoruba, Asante Twi, and Akuapem Twi all
from BibleTTS~\cite{meyer2022bibletts}. 
From VoxLingua107, it contains Hungarian and Finnish for Finno-Ugric languages; Maori and Hawaiian for Austronesian languages; Georgian for Caucasian languages; Nepali, Hindi, and Urdu for Indo-Aryan languages; and Thai for Tai languages.

Within the dataset, there is one folder for each language denoted by its ISO 639-2 code, and within a folder there are a
number of \path{wav} files and their corresponding utterance transcription \path{txt} files sharing the same name. 
The naming convention for every file follows the format
\path{<language-code>_<source-dataset>_<sex>_<speaker-id>_<index>.wav}
(or \path{.txt}). The sex and speaker-id fields are replaced with a
``\path{u}'' if unavailable.

\section{Method}
\subsection{In-Set and Out-of-Set}
Since the goal of this work was to not only identify known languages, but also to detect, learn, and identify unknown languages, we split our dataset into two categories: in-set and out-of-set. 
The CU MultiLang Dataset was split into 32 in-set
languages and 19 out-of-set languages while ensuring that both categories remained diverse and
encompassing. See Table \ref{table:cumultilang} for the in-set and
out-of-set breakdown. 

\begin{table}
\caption{In-set and Out-of-set languages}
  \centering
\begin{center}
\begin{tabular}{ll|ll}
    \toprule
    \multicolumn{2}{c}{In-Set}  &  \multicolumn{2}{c}{Out-of-Set} \\
    \midrule
    Arabic      &  Kashmiri &                Akuapem Twi & Romanian\\
    Bengali     &  Korean    &        Albanian & Ukrainian\\
    Catalan     &  Lingala   &          Armenian & Uyghur\\
    English     &  Mandarin  &              Asante Twi  \\
    Ewe         &  Maori  &             Bulgarian \\
    French      &  Pashto  &               Burmese  \\
    Georgian    & Russian   &              Croatian  \\
    German      &  Spanish  &             Dutch \\
    Greek       &  Swedish  &               Finnish  \\
    Hausa       &  Tamil    &              Hebrew\\
    Hawaiian    &  Telugu &                Iban     \\
    Hindi       &   Thai    &   Japanese \\
    Hungarian   & Tibetan & Malayalam \\
    Icelandic   & Turkish   &   Nepali \\
    Italian     & Urdu &    Norwegian \\
    Javanese        & Yoruba    &   Persian 
\end{tabular} 
\end{center}
\label{table:cumultilang}
\end{table} \ \\


\subsection{Data Standardization}
All utterances were standardized by batching the data into fixed-length segments which greatly improved the stability and speed during training. 
To do identify the optimal segment length, the audio data was split into fixed-length segments of 2, 3, 4, and 5 seconds for experimentation. Five hours of audio were taken from each language and the TDNN was trained for 15 epochs using a batch size of 512.  
Four seconds yielded the highest validation accuracy, 
so all utterance samples were standardized as a series of several four-second segments.

\subsection{Train and Test Sets}
The training data for the TDNN
is 95\% of the total duration of each in-set language with
a 90/10 split for train/validation. For the initial fit of our
LDA and pLDA on our out-of-set languages, 80\% of the total duration
of each out-of-set language was used.
During training, we utilized a speaker leave-out technique where possible (and later used left-out speakers for evaluation), in order to ensure the system was learning language-specific features instead of speaker-specific features.

The remaining 5\% of each in-set language and 20\% of each
out-of-set language was used as the test set.

\subsection{Feature Extraction}

Kaldi~\cite{r-m:povey-2011} was used to extract MFCC features from our utterances. 
We then
concatenated additional pitch information, as it proved to add
meaningful information in ``Modernizing Open-Set Speech Language Identification.'' This results in a final
feature embedding for each utterance that is a series of 16-dimension vectors (one per time slice).

\subsection{System Architecture}
Final feature embeddings are passed through a TDNN which was architected and trained from
scratch without any pre-trained models or transfer learning. The TDNN produces two outputs. The first is the output of the final layer of the TDNN: a
32-dimension softmax output. This TDNN output for a sample is averaged
over all time-slices to get a single 32-dimension softmax output. It
is then passed through a threshold function which observes the softmax
probabilities to determine whether the sample belongs to an in-set
language or an out-of-set language. If
the sample is determined to be an in-set language, the softmax output
is used to make a language prediction with highest probability. This
was the extent of the system in ``Modernizing Open-Set Speech Language Identification.''

\begin{figure}[t]
\centerline{\includegraphics[scale=0.38]{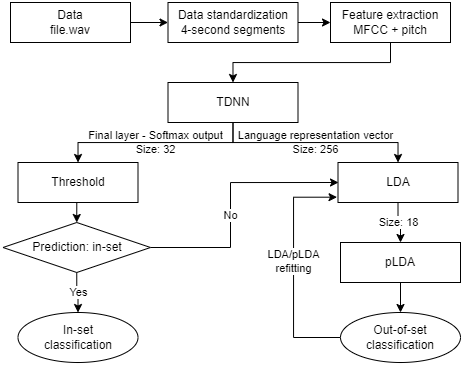}}
\caption{System architecture}
\label{fig:sys-arch}
\end{figure}

The second output of the TDNN is a 256-dimension language
representation vector. This is the output of the TDNN when peeling
back its last two layers: the softmax layer and the 32-dimension
penultimate layer. If the language is rejected and classified as
out-of-set, we then concatenate all language representation vectors
from each time slice in the sample to preserve as much data as we can,
and pass the result through an LDA in order to reduce the dimension of
our representation vectors down to 18, performing both
dimensionality and correlation reduction of the features. Finally, the
18-dimension vectors are passed through a pLDA to give us a classification of our out-of-set language. If the pLDA classifier cannot
confidently predict the out-of-set language, then we potentially have
a new language to fit into our system. To do so, we simply re-fit the
LDA and pLDA components given the new data and a new out-of-set
language class label to match it. Thus, our system can
now recognize a new language without having to undergo the
heavyweight task of retraining or fine-tuning the TDNN.
See Fig. \ref{fig:sys-arch} for the
full architecture of our open-set spoken LID system.

\section{Accessibility and Accuracy}
To make sure that any 32GB memory system with a consumer-level GPU could train, fine-tune, or perform inference with our spoken LID system, we restricted the entirety of our work to a machine with 30GB of memory and a single NVIDIA Tesla K80 GPU. 

\subsection{TDNN Optimization and Size Reduction}
We first reduced the size of the TDNN while retaining or
increasing accuracy to make training the model and using it for inference faster.

Using the same TDNN model from ``Modernizing Open-Set Speech Language Identification,'' the output dimension of the language representation
vectors was 1500, which spread relevant information thin and made the process of fitting our LDA component require far too much
memory. 
We reduced the language representation layer down to a size of 256, giving the model a total parameter count of only 556,896 while also increasing in-set identification accuracy and out-of-set detection accuracy.
A primary factor for the observed accuracy improvements was the
utilization of the AdamW optimizer~\cite{adamw}, which improved convergence
compared to the previously used SGD optimizer. 

The final TDNN model architecture was five layers of size 256 followed by one layer of size 32. All six layers used dilation and stride of 1, but the first three layers used context size of 3 and the last three layers used context size of 1. Finally, a softmax output layer was added at the end of the TDNN model. Additionally, batch normalization was applied after every layer to improve generalization, and help reduce noise from other features of the audio.

\subsection{Data Pipeline and Ensemble Approach}
We kept a tight constraint on our data pipeline's
memory usage. Despite reducing our language representation vectors
down to size 256, we could not load all of them at once to fit the LDA and pLDA all at once. 
In order to make sure that less than 32GB of data were loaded at any given time, the 256-length vectors were loaded in batches, at approximately 4,000 four-second segments' worth at a time. 

\begin{figure}[t]
\centerline{\includegraphics[scale=0.38]{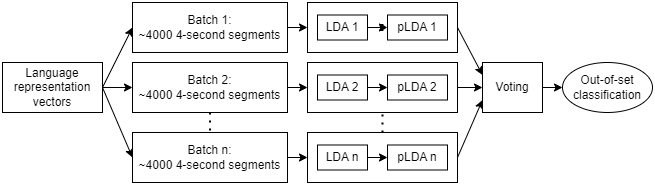}}
\caption{LDA and pLDA ensemble architecture}
\label{fig:ldaplda-arch}
\end{figure}

We then implemented an ensemble algorithm for the LDA
and pLDA layers. Ensuring that each batch of 4,000 four-second
segments was well-diversified, we fit a unique LDA and pLDA
for each batch. To use this for testing and inference, we pass a sample through all of
our LDA and pLDA pairs, taking the most common prediction with
mean confidence to be our final output. This allowed us to circumvent
the memory restrictions of fitting an LDA and pLDA while keeping our
data pipeline constrained. 
See Fig. \ref{fig:ldaplda-arch} for a
diagram of the data pipeline and ensemble architecture.

\section{Results}



\subsection{In-set Accuracy}

\begin{figure}[t]
\centerline{\includegraphics[scale=0.4]{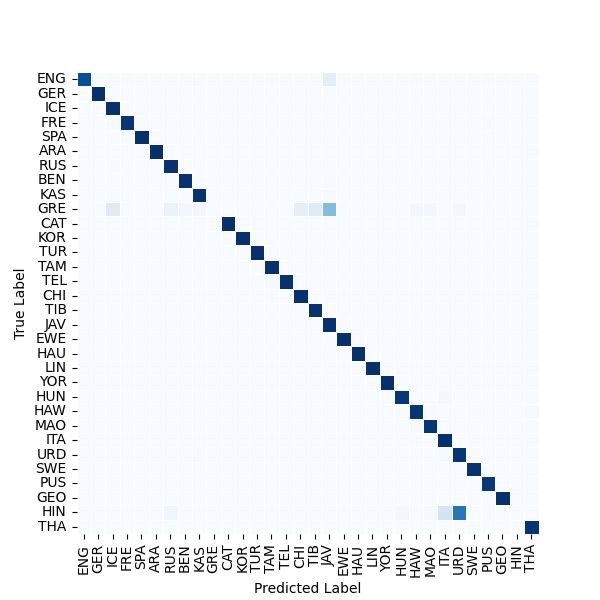}}
\caption{In-set language confusion matrix}
\label{fig:cmis}
\end{figure}

On in-set languages, our system achieves 91.76\% accuracy via the TDNN model. When observing top-N accuracies, we found that TDNN achieves 94.20\% top-2 accuracy, 95.05\% top-3 accuracy, 95.80\% top-4 accuracy, and 96.18\% top-5 accuracy. Observing the confusion matrix for in-set languages in Fig. \ref{fig:cmis}, we can see that the TDNN often confuses Hindi with Urdu, English with Javanese, and Greek with Javanese. 

When we consider how our system will be used with confidence thresholding, a higher threshold results in our system labeling fewer and fewer inputs as in-set languages. In effect, confidence thresholding ensures that the system is more confident in its in-set labelling and results in a higher in-set classification accuracy. Of the inputs correctly labeled as in-set with a threshold of 0.65, the system accurately assigns the correctly language with 98\% accuracy. 

\subsection{Out-of-set Accuracy}

\begin{figure}[t]
\centerline{\includegraphics[scale=0.4]{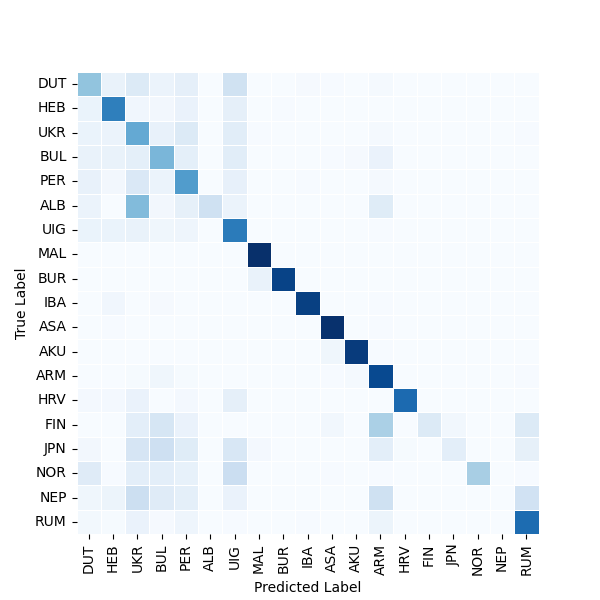}}
\caption{Out-of-set language confusion matrix}
\label{fig:cmoos}
\end{figure}

On out-of-set languages, our system achieves 72.93\% accuracy via the LDA and pLDA layers. Observing the confusion matrix for in-set languages in Fig. \ref{fig:cmoos}, we can see that the LDA and pLDA layers often confuse Albanian with Ukrainian, Finnish with Armenian, and Nepali with Romanian. 

\subsection{Optimal Threshold}

Choosing the optimal threshold for an open-set language identification system is difficult because there is no threshold that fits all use cases. Depending on the user, there are various operating conditions that constitute different choices of confidence thresholds. 

\begin{figure}[t]
\centerline{\includegraphics[scale=0.4]{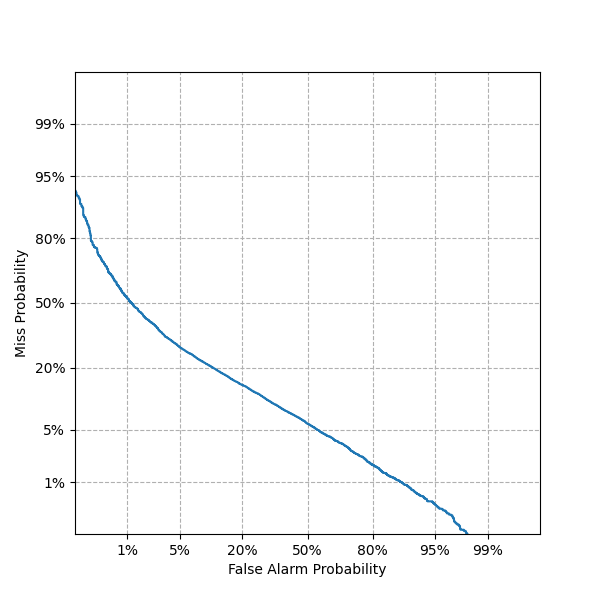}}
\caption{DET curve}
\label{fig:det}
\end{figure}

One way of choosing a threshold is to analyze our system as a whole and thinking of the open-set spoken language identification task as a binary classification problem. We frame the question as such: is the given input from an in-set language or an out-of-set language? Fig. \ref{fig:det} plots miss probability, the rate of in-set samples incorrectly labeled as out-of-set, on the vertical axis and false alarm probability, the rate of out-of-set samples incorrectly labeled as in-set, on the horizontal axis. This DET curve~\cite{Martin1997TheDC} thus represents the performance of our spoken language identification system and shows the trade-off of error types. We can report the point of equal error rate on the DET curve: both probabilities are 19\% at a threshold of 0.65. 

The second way of choosing a threshold is to measure total accuracy of our system when using various confidence thresholds. For all samples in our test split, how often did the system label an input with the correct language class? When using a threshold of 0, the system achieves 34.49\% because all samples are labeled with in-set languages, meaning all out-of-set samples are mislabeled. When using a threshold of 1, the system achieves 45.53\% accuracy because all samples are labeled with out-of-set languages, meaning all in-set samples are being mislabeled. With a threshold of 0.81, the system achieves its maximum total accuracy of 69.80\%; thus, 0.81 may be another optimal threshold. However, it should be noted that the total accuracy numbers are highly reflective of the proportion of in-set and out-of-set samples in the test split.  

\begin{figure}[t]
\centerline{\includegraphics[scale=0.4]{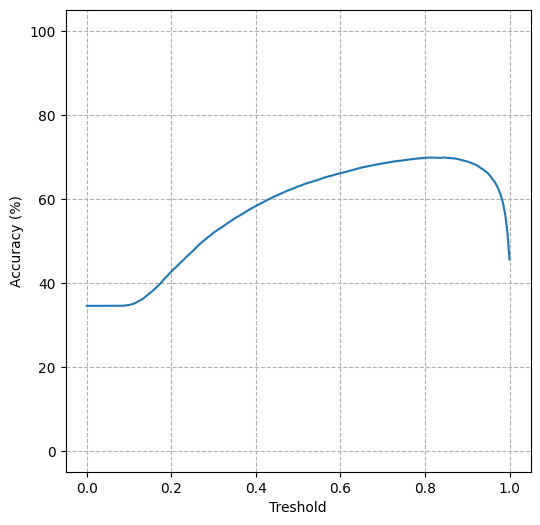}}
\caption{Total accuracy}
\label{fig:tacc}
\end{figure}

\section{Conclusion and Future Work}
The results show promise in achieving our goal of creating an
accessible and robust spoken language identification system, that can both identify known languages with high accuracy and also recognize unknown languages. 
Furthermore, the creation of the
open-source CU MultiLang Dataset will hopefully prove to
be useful for developers working on related spoken language
problems. 

We could improve our open-set spoken LID system with a more diverse array
of thresholding mechanisms. Dynamic solutions such as having a threshold for each language, may yield significantly improved
results. Additionally, more can be done with respect to feature
abstraction and embedding. Further experimentation with respect to incorporation of other spectral features and i-vectors
may help in increasing component accuracies.

Finally, to test the system in real-world scenarios, we may build an application that allows users to interface with the system by recording audio clips of themselves. We
would test our in-set prediction accuracy and also see how well the application recognizes and adapts to new out-of-set languages. The application may also
provide data that, with user permission, could be added to the CU
MultiLang dataset.

\bibliographystyle{IEEEtran}
\bibliography{ms}

\end{document}